\pdfoutput=1
\documentclass[11pt]{article}

\usepackage{acl}

\usepackage{times}
\usepackage{latexsym}

\usepackage[T1]{fontenc}
\usepackage[utf8]{inputenc}

\usepackage{microtype}

\usepackage{inconsolata}

\usepackage{graphicx}
\usepackage{tcolorbox}
\usepackage{xcolor}
\usepackage{colortbl}

\usepackage{booktabs}

\definecolor{lightlightgray}{rgb}{0.9, 0.9, 0.9}
\newcolumntype{a}{>{\em\columncolor{lightlightgray}}c}

\title{Explicit Inductive Inference using Large Language Models}

\author{Tianyang Liu \quad Tianyi Li \quad Liang Cheng \quad Mark Steedman \\
  University of Edinburgh \\
  \texttt{T.Liu-47@sms.ed.ac.uk \quad tianyi.li@ed.ac.uk} \\
  \texttt{L.Cheng-13@sms.ed.ac.uk \quad m.steedman@ed.ac.uk}}

\begin{document}
\maketitle
\begin{abstract}
Large Language Models (LLMs) are reported to hold undesirable attestation bias on inference tasks: when asked to predict if a premise $P$ entails a hypothesis $H$, instead of considering $H$'s conditional truthfulness entailed by $P$, LLMs tend to use the out-of-context truth label of $H$ as a fragile proxy. In this paper, we propose a pipeline that exploits this bias to do explicit inductive inference. Our pipeline uses an LLM to transform a premise into a set of attested alternatives, and then aggregate answers of the derived new entailment inquiries to support the original inference prediction. On a directional predicate entailment benchmark, we demonstrate that by applying this simple pipeline, we can improve the overall performance of LLMs on inference and substantially alleviate the impact of their attestation bias.\footnote{Codes and data of this paper are available at \url{https://github.com/waterltyang/EIDI}}
\end{abstract}

\section{Introduction}
\label{sec:intro}
% Large Language Models' (LLMs') outstanding performance on various natural language tasks recently attracts a large amount of research. One essential aspect of it is to measure LLMs' potential to do reasoning. Evidences are showing that LLMs have gained reasoning ability to a noticeable extent, yet their limits and the underlying mechanism are still to be revealed. LLMs encode massive facts during pre-training, which makes it difficult to probe if they are faking reasoning abilities from their advantage of memorization. As a result, besides inspecting LLMs' behaviour in complex reasoning scenarios, more attention is also drawn to testing LLMs with basic inference task
Large Language Models (LLMs) are claimed to possess \textit{implicit} inductive reasoning ability through pre-training: from the massive examples they memorized, they draw inference rules and encode them latently so that they can apply these rules to do reasoning at test time.

However, recently \citet{mckenna-etal-2023-sources} has pointed out that LLMs are severely affected by an attestation bias when performing inference tasks. Given the question of whether premise $P$ entails hypothesis $H$ with few-shot examples, an LLM's prediction is deeply bound to the hypothesis' out-of-context truthfulness, instead of its conditional truthfulness entailed by the premise. When the hypothesis $H$ is attested in an LLM's world knowledge (the LLM believes $H$ to be true), the LLM is likely to predict the entailment to be true, regardless of the premise. As a result, LLMs suffer a significant performance drop when the entailment labels disagree with the attestation label of hypothesis $H$.
% , compared to cases where these two labels agree. 

Although this is a severe problem limiting LLMs' performance on non-attested inferences, we argue that with careful design, this bias can instead be \textbf{exploited} to improve LLM performance on inference tasks. We notice a statistically true conclusion: Given an entailment inquiry $P \models H$, the attestation bias is harmful only when the premise $P$ is not attested. If we control $P$ to always be attested, then $P \models H$ will naturally share the same truth label with $H$ on a distributional basis, which dissolves the negative effects of LLMs' attestation bias. 

\begin{figure}[t]
  \includegraphics[width=1\linewidth]{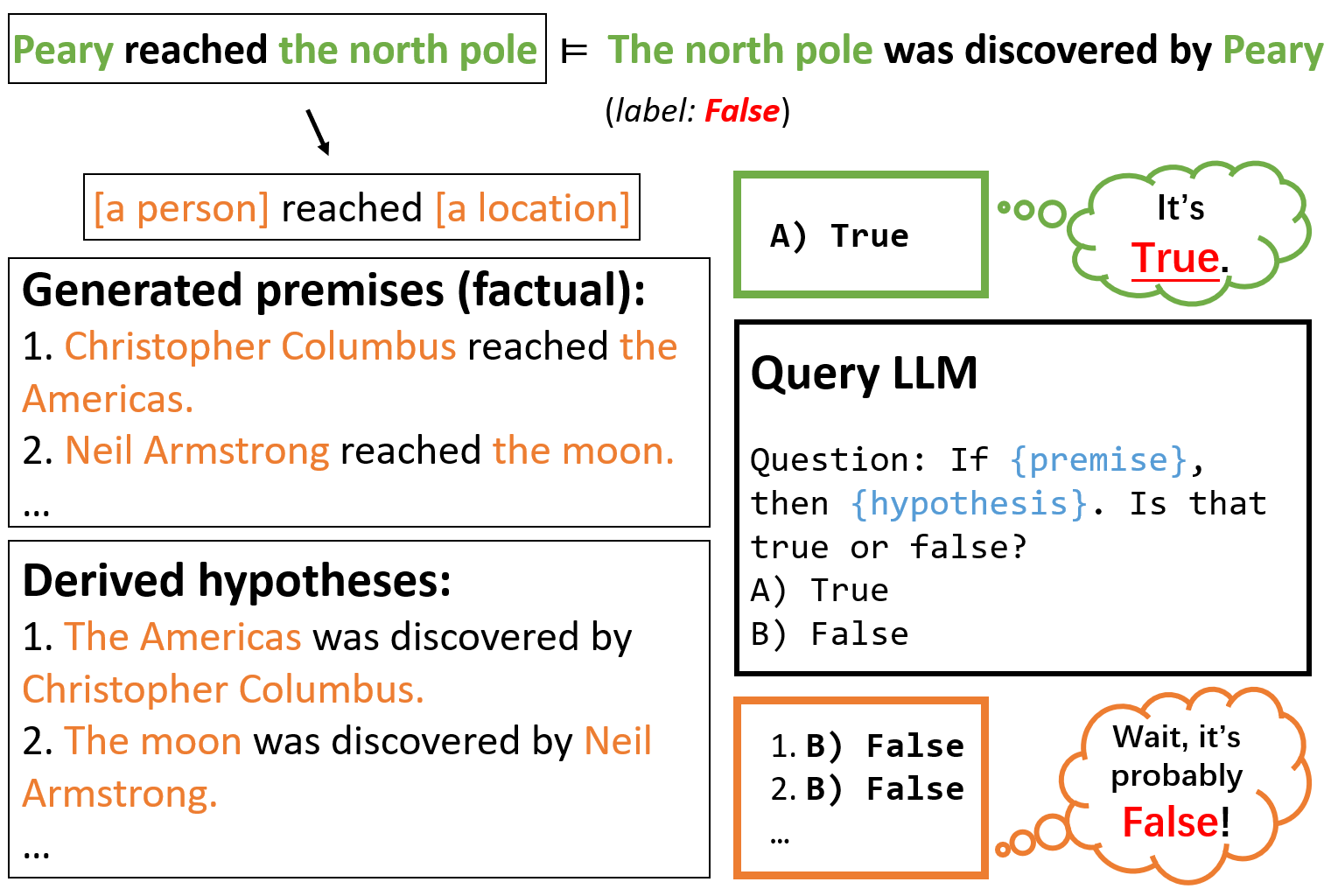} 
  \caption {An example of the explicit inductive inference pipeline. While direct entailment inquiry gets a wrong answer, it can be corrected by reasoning on more alternative examples. }
  \label{workflow}
  \vspace{-0.2in}
\end{figure}

Applying this idea, we propose a simple yet effective Explicit Inductive Inference pipeline with LLMs. As illustrated in Figure \ref{workflow}, the core idea is to transform a premise into a set of attested alternatives by replacing the arguments, and to aggregate the LLM's predictions on these derived inquiries to support answering the original question. 

We test our pipeline with two LLMs on Levy/Holt \cite{levy-dagan-2016-annotating, holt2019probabilistic}, a difficult directional predicate inference dataset, and further analyze the influence of our pipeline against the models' attestation bias. The results show that our pipeline can improve not only LLM's overall performance on predicate inference, but also their robustness against the attestation bias.

To summarize our contribution, we propose an easy-to-use inference pipeline that \textbf{1)} improves LLMs' performance on predicate inference, \textbf{2)} substantially alleviates negative effects of the LLMs' attestation bias, and \textbf{3)} uses LLMs' own generation capability without requiring external knowledge.

\section{Related Work}

LLMs accumulate a bias towards factual knowledge by encoding massive facts during pre-training \cite{roberts-etal-2020-much, carlini2022quantifying, yan-etal-2022-robustness}. Recently, \citet{mckenna-etal-2023-sources} pointed out that LLMs suffer from an attestation bias on inference tasks as a result. Note that the effect of attestation bias is similar to that of the hypothesis-only baseline \cite{poliak-etal-2018-hypothesis}, but while the former is a bias from pre-training, the latter originates from dataset artifacts in supervised learning.

In other tasks, previous work has mitigated the bias towards attestation by introducing counterfactual examples \cite{wang-etal-2022-rely, zhou-etal-2023-context, wang-etal-2023-causal} or replacing argument entities with their type labels \cite{zhou2024conceptual}. In this paper, we go one step further to show that in an inference task, we should instead encourage the models to generate factual alternative examples. 
% With awareness of the hypothesis only trap \citet{poliak-etal-2018-hypothesis}, we use the argument substitution of the premises to instantiate the hypothesis triple to guarantee their dependency.

The idea of aggregating multiple versions of LLMs' outputs has been explored in prior work. \citet{wang2022self} encourage LLMs to generate multiple reasoning paths for one question, while \citet{zhou-etal-2022-prompt} embody one question with multiple prompts. In contrast, our method creates semantically different alternative questions, which serve as extra evidence for an original inquiry.

%% On evaluation end, \citet{poliak-etal-2018-hypothesis} exposed the weakness of majority of inference datasets upon which a baseline model that only looks at the hypothesis can have decent performances. This further motivate us to design experiments to  

% The ability of LLMs to maintain the same response when a question is presented in various forms is sometimes discussed under the topic of self-consistency.

% There are other works reporting performance improvement on other tasks by aggregating LLMs' responses to different question forms \cite{wang2022self, zhou-etal-2022-prompt}, but these works approach this problem in a prompt engineering way, while our work focuses more on the semantic level of the predicate inference task.

\section{Explicit Inductive Inference}

\subsection{Task and Definition}

The task of this work is to determine the entailment relation between two binary predicates where both predicates are contextualized with the same pair of entities. The input will be in the form of two triples $(s, p, o) - (s, h, o)$ where $s$ is the subject entity, $o$ is the object entity, $p$ is the premise predicate, and $h$ is the hypothesis predicate. There are also cases in the form of $(s, p, o) - (o, h, s)$ where the two entities are swapped in position like the example in Figure \ref{workflow}. Without loss of generality, we describe our method with inputs in the former format.

The goal is to predict whether the premise triple entails the hypothesis triple, namely the truth label of $(s, p, o) \models (s, h, o)$. To use an LLM to predict entailments, each input triple pair will be wrapped into a prompt. We mark them as $Q[(s, p, o) \models (s, h, o)]$ and call them entailment inquiries.

\subsection{Exploit the Attestation Bias}

As stated in Section \ref{sec:intro}, the attestation bias of LLMs can be less detrimental if the premise $P$ is attested in an entailment inquiry, because the truth label of $P \models H$ would likely be the same as the attestation label of $H$. Besides this, two more insights are guiding our pipeline design:

1) The label of a predicate entailment inquiry does not change when the argument entities are replaced, as long as the substitution entities keep the same semantic type labels.

2) Factual $\neq$ Attested. Factual knowledge from external sources may not be confirmed by LLMs for being longtail, absent in pre-training data, or conflicted with out-of-date records. Facts generated by LLMs, on the other hand, are highly likely to be recognizable by themselves. Even hallucinated generations are acceptable since they are still attested if not factual.

Based on these understandings, we propose the \textbf{E}xplicit \textbf{I}n\textbf{D}uctive \textbf{I}nference (EIDI) pipeline. Given an entailment inquiry $P \models H$, our EIDI pipeline first transforms $P$ into a set of different attested premises $P'$s by replacing the arguments of $P$. Then the corresponding set of $H's$ is derived, so that we now have a list of alternative inquiries $P' \models H'$. Finally, we explicitly do an inductive inference on these new inquiries by drawing a concluded entailment prediction from an LLM's answers to these alternative inquiries.

It is worth mentioning that given $P$ is true, logically,  $H$ is always true if $P \models H$ but not vice versa. We can only statistically conjecture $P \models H$ if we observe a high probability of $H$ being true (predicted by the LLM according to the bias). 
Therefore, we encourage the transformation module to generate a variety of different alternative premise triples, so that a more reliable conclusion can be drawn when we aggregate the predictions.

% In the remainder of this section, we introduce the three modules in the EIDI pipeline in detail.

\subsection{Explicit Inductive Inference Pipeline}
\label{sec:pipeline}
% We now introduce the three modules of the EIDI pipeline in detail.

\paragraph{Typing} While the label of (medicine X, \textit{kills}, disease Y) $\models$ (medicine X, \textit{is a cure of}, disease Y) is \textit{True}, one can not therefore deduce that (Person X, \textit{kills}, animal Y) $\models$ (Person X, \textit{is a cure of}, Animal Y). To prevent these errors incited by the ambiguity of predicates, for each premise triple $(s, p, o)$, we query the LLM to obtain the entity type label of the subject and object $t_s$ and $t_o$. Here we do not predefine a vocabulary for possible type labels since the purpose is only to disambiguate.

\paragraph{Transformation} With these assigned type labels we query the LLM to generate alternative arguments for the premise predicate. From one typed premise triple $(s, t_s, p, o, t_o)$, we encourage the LLM to generate a list of new attested triples $(s_1, p, o_1), ..., (s_n, p, o_n)$ where the substitution entities keep the original types, i.e. any $s_i$ still has type $t_s$ and any $o_i$ still has type $t_o$. These $n$ new premise triples will then be expanded to $n$ new entailment inqueries $Q[(s_1, p, o_1) \models (s_1, h, o_1)], ..., Q[(s_n, p, o_n) \models (s_n, h, o_n)]$.

\paragraph{Prediction} At this point, we input each derived entailment inquiry $Q[(s_i, p, o_i) \models (s_i, h, o_i)]$ to the LLM to get their response and corresponding probability score. Then we take the average score of these $n$ different scores as our explicit inductive score for the original entailment inquiry.

\section{Experimental Setup}

% We now introduce the experimental details of how we test our methods.

\subsection{Datasets}

We test our pipeline on the Levy/Holt dataset \cite{levy-dagan-2016-annotating, holt2019probabilistic}, a predicate entailment dataset where each entry consists of two triples in the form of $(s, p, o) - (s, h, o)$ or $(s, p, o) - (o, h, s)$, and a following label shows whether the premise triple entails the hypothesis triple. 
We use the directional portion of this dataset following prior work \cite{mckenna-etal-2023-smoothing, chen-etal-2022-entailment, li-etal-2022-language}, as it is a challenging test focused on the understanding of entailment beyond bi-directional similarity.

Following \citet{mckenna-etal-2023-sources}, we further analyze how the LLMs' attestation bias is digested in our method. We split the Levy/Holt dataset according to whether the label of $P \models H$ agrees with the attestation label (obtained by querying the LLM) of $H$ for each entry. For the 1784 entries in the full directional test set, this yields an attestation-consistent subset of 956 entries and an attestation-adversarial subset of 828 entries.\footnote{The substantial size of the attestation-adversarial subset demonstrates the detrimental effect of attestation bias in real datasets.} We report results on both the directional test set and its two subsets in Section \ref{sec:results}.

\subsection{LLMs}

We test our method with two LLMs, GPT-3.5 and Llama3. GPT-3.5 \cite{gpt3.5} is a set of powerful closed-source commercial LLMs. We choose the GPT-3.5-Turbo version for its widespread use in the research community. Llama3 \cite{llama3} is a SOTA open-source LLM, where we choose the largest Llama3-70B-instruct version for its optimized capacity. Throughout our experiments, we use greedy decoding for reproducible results.

Our pilot studies on the development set indicate that adding few-shot examples in the prediction module may add extra bias to the model, and therefore introduce unnecessary considerations on finding proper examples under each setting. Hence we choose zero-shot prompts for the prediction module and one-shot prompts for the transformation module where the only example is the original premise. Discussion on prompt selection and a list of all prompts we use are included in Appendix \ref{apd:a}.

\subsection{Baselines and Metric}

We compare EIDI against two baselines. We construct MCQ$_{entity}$ baseline by directly wrapping the original premise and hypothesis with the Multipe-Choice Question prompt used in our prediction module, and passing it to the LLM to get an entailment prediction. MCQ$_{type}$ baseline is set up in the same way where the only difference is that we first replace the arguments of the predicates by their entity types. To 
% make our results comparable to previous work
keep ourselves aligned with previous work, we use the 48 FIGER types \cite{ling2012fine} as in \citet{mckenna-etal-2023-sources} for this measure, instead of the LLM-generated types in Section \ref{sec:pipeline}.

We draw the precision-recall curve for EIDI and each baseline by inspecting the final output token probability of the model's response. As a result of the multiple-choice prompt design, returned answers always start with a choice mark where A is for entailment and B is for non-entailment. For baseline methods, we score that one token's probability. For our EIDI pipeline, we calculate the average score of the $k$ output tokens' probabilities.

Following \citet{li-etal-2022-language, mckenna-etal-2023-sources}, we calculate the normalized area-under-curve (AUC$_{norm}$) as an indicator of the model's performance. This measure describes how much better a model is over a degenerate baseline returning positive answers to every data entry.

\section{Results and Discussion}
\label{sec:results}

\begin{table}[]
\centering
\begin{tabular}{@{}lcc@{}}
\toprule
                & \multicolumn{2}{c}{\textbf{Model}} \\ \cmidrule(l){2-3} 
\textbf{Pipeline}& GPT-3.5              & Llama3               \\ \midrule
MCQ$_{entity}$    & 23.85                & 36.66                \\
MCQ$_{type}$      & 25.88                & 35.13                \\
EIDI$_{all}$      & \textbf{35.52}       & \textbf{50.89}       \\ \midrule
EIDI$_1$          & 31.16                & 41.85                \\
EIDI$_2$          & 32.10                & 46.75                \\
EIDI$_5$          & 33.41                & 49.61                \\ \bottomrule
\end{tabular}
\caption{\label{overall_results}
Overall normalized Area-Under-the-Curve (\%) of our EIDI pipeline and the two baselines on the full Levy/Holt directional test set. EIDI$_i$ inspects only $i$ alternative inquiries, and EIDI$_{all}$ considers all examples obtained in the transformation step.
}
\vspace{-0.2in}
\end{table}

\subsection{Overall performance}

Table \ref{overall_results} shows the performance of each model on the directional Levy/Holt test set. With both LLMs, our EIDI$_{all}$ pipeline gains a significant improvement over the two baseline methods.

The typical value of the size of total generated examples $n$ is $10$ for the EIDI$_{all}$ setting. It can be observed that the performance of EIDI$_i$ steadily increases along with $i$, confirming our hypothesis that with attested $P'$s, the more cases of alternative $P' \models H'$ generated, the more reliable our pipeline is. The complete results of all EIDI$_i$ settings are shown in Appendix \ref{apd:all}.

An interesting observation lies between the performance of the EIDI$_1$ setting and the baselines, which shows that replacing the original inquiry with even one self-generated example can improve the LLMs' predicate inference performance. The difference between EIDI$_1$ and MCQ$_{type}$ baseline also highlights the importance of instantiating attested triples. Since the effect of the attestation bias is already excluded from the results of the MCQ$_{type}$, this proves that the EIDI pipeline is taking advantage of further exploiting the bias.

\begin{table}[h]
\centering
\begin{tabular}{@{}llcca}
\toprule
\textbf{Model} & \textbf{Pipeline}                     & \textit{cons.} & \textit{adv.}              & \textit{\textbf{diff.}} \\ \midrule
GPT-3.5 & \multicolumn{1}{l|}{MCQ$_{entity}$}         & 82.04 & \multicolumn{1}{c|}{0.00}  & -82.04 \\
        & \multicolumn{1}{l|}{MCQ$_{type}$}          & 69.40          & \multicolumn{1}{c|}{0.48}  & -68.92 \\
               & \multicolumn{1}{l|}{EIDI$_{all}$} & 56.14                   & \multicolumn{1}{c|}{9.97}  & \textbf{-46.17}         \\
        & \multicolumn{1}{l|}{EIDI$_1$} & 53.73          & \multicolumn{1}{c|}{8.95}  & \textbf{-44.78} \\ \midrule
Llama3  & \multicolumn{1}{l|}{MCQ$_{entity}$}         & 81.08 & \multicolumn{1}{c|}{0.01}  & -81.07 \\
        & \multicolumn{1}{l|}{MCQ$_{type}$}          & 70.25          & \multicolumn{1}{c|}{2.41}  & -67.84 \\
               & \multicolumn{1}{l|}{EIDI$_{all}$} & 69.59                   & \multicolumn{1}{c|}{23.83} & \textbf{-45.76}         \\
        & \multicolumn{1}{l|}{EIDI$_1$} & 63.98          & \multicolumn{1}{c|}{15.66} & -48.32 \\ \bottomrule
\end{tabular}
\caption{
AUC$_{norm}$ (\%) on the attestation-bias-split datasets. The \textbf{\textit{diff.}} column marks the difference between results on the attestation-consistent (\textit{cons.}) and attestation-adversarial (\textit{adv.}) subsets. 
}
\vspace{-0.2in}
\label{tab:att}
\end{table}

\subsection{Against the bias}

Table \ref{tab:att} compares the performance of each method on attestation-consistent (\textit{cons.}) and attestation-adversarial (\textit{adv.}) subsets. Measured by the difference of AUC$_{norm}$ between the two subsets, our pipeline reduces the effect of LLMs' attestation bias by over 20\% from the MCQ$_{type}$ baseline, and over 35\% from the MCQ$_{entity}$ baseline in average.

With both LLMs, we observe an AUC$_{norm}$ of near 0\% in the two baseline settings, demonstrating the extreme inability of the LLMs to capture the essential entailment signal against the attestation bias in a zero-shot setting. 

Interesting results appear again under the EIDI$_1$ setting. On GPT-3.5-turbo, it slightly outperforms the EIDI$_{all}$ setting. But this only happens because EIDI$_{all}$ setting is doing better on the attestation-consistent subset, which implies that EIDI$_{all}$ setting is still the choice for best performance, while EIDI$_1$ is also a strong candidate for scenarios with limited compute.

These results suggest that our pipeline can be used to improve LLMs' general inference performance, and especially in attestation-adversarial scenarios, e.g. \textit{If lions are fed on hay, then lions eat hay. } As a replacement to LLM's direct inference prediction, EIDI pipeline can be easily plugged into frameworks with LLMs to do various downstream tasks like question answering and KG completion.

\section{Conclusions}

We propose an explicit inductive pipeline exploiting the attestation bias of LLMs to do more robust predicate inference. With experiments on the directional Levy/Holt dataset and its attestation-bias-split subsets, we have shown that our baseline gains a significant improvement over LLM's primary inference performance, and substantially reduces the performance loss caused by LLMs' attestation bias. 

Without external knowledge, EIDI use LLMs' own generation to exploit their attestation bias. Our results suggest that although biases of LLMs are usually undesirable obstacles, in some scenarios they may be tapped for good with careful design. We advocate for similar ideas to be applied to other tasks to improve LLM performance in future work.

\section*{Limitations}

In this paper, we demonstrated the performance of our pipeline by comparing it to two baselines. Although we intend to exclude prompt engineering factors from our analysis, it has been widely accepted that including few-shot examples and other prompting techniques can guide LLMs to output better answers. Therefore there could be further studies on evaluating the effects of using different prompts in the EIDI pipeline.

Generating alternative inquiries and respectively doing inferences on them can be computationally expensive compared to only one determination in baseline settings. As a result, downstream applications may find a trade-off between computational efficiency and better inference performance.

We also tested our pipeline against the frequency bias that \citet{mckenna-etal-2023-sources} pointed out, and the results show that the EIDI pipeline amplifies this bias compared to the baselines due to its choice of popular entities. We argue that this reaffirms the challenge in achieving Pareto improvements on LLM robustness against biases, and leave those results and discussions to Appendix \ref{apd:c}.

\bibliography{anthology, custom}
% \bibliography{custom}

\clearpage
\appendix

\section{Prompts Selection}
\label{apd:a}

Here we list all the prompts that we use in our experiments.

\paragraph{Typing} The purpose of this module is only to disambiguate the predicates, therefore no vocabulary of allowed type labels is predefined.
\begin{quote}
    Type the entities in the following triples: 

    Hitler | was born in | Poland -> a person | was born in | a country

    Hogs | eats | Corn -> an animal | eats | a food

    Aspirin | may reduce the risk of | Cancer -> a medicine | may reduce the risk of | a disease

    \{$s$\} | \{$p$\} | \{$o$\} -> 
\end{quote}

\paragraph{Transformation} Although we use the word 'fact', the generated triples are always attested rather than factual.
\begin{quote}
    Write \{$n$ + 1\} facts in the form of " \{$t_s$\} | \{$p$\} | \{$t_o$\}."
    
    - \{$s$\} | \{$p$\} | \{$o$\}. 
    
    -
\end{quote}

\paragraph{Prediction} This is also used for the two baselines. 

\begin{quote}
    Question:If \{$s$\} \{$p$\} \{$o$\}, then \{$s$\} \{$h$\} \{$o$\}. Is that true or false?
    
    Choices:
    
    A) True
    
    B) False
    
    Answer:
\end{quote}

For prediction module, when an instruction is required, we use the following one:

\begin{quote}
    Only return one mark A, B or C to answer the question.
\end{quote}

\section{Results on all EIDI$_i$ Settings}
\label{apd:all}

Table \ref{all_setting} shows the performance of all EIDI$_i$ settings. Best performences are reached when all transformed alternative inquiries are considered.

\begin{table}[h]
\centering
\begin{tabular}{@{}lcc@{}}
\toprule
                & \multicolumn{2}{c}{\textbf{Model}} \\ \cmidrule(l){2-3} 
\textbf{Pipeline}& GPT-3.5              & Llama3               \\ \midrule
MCQ$_{entity}$    & 23.85                & 36.66                \\
MCQ$_{type}$      & 25.88                & 35.13                \\
\midrule
EIDI$_1$          & 31.16                & 41.85                \\
EIDI$_2$          & 32.10                & 46.75                \\
EIDI$_3$          & 31.47                & 47.52\\
EIDI$_4$          & 32.05                & 48.60\\
EIDI$_5$          & 33.54                & 49.61\\
EIDI$_6$          & 33.41                & 50.42\\
EIDI$_7$          & 34.68                & 50.13\\
EIDI$_8$          & 34.76                & 50.36\\
EIDI$_9$          & 35.28                & 50.39\\
EIDI$_{10}$       & \textbf{35.52}                & 50.01\\
EIDI$_{11}$       & -                    & 50.52\\
EIDI$_{12}$       & -                    & \textbf{50.89}\\
\bottomrule
\end{tabular}
\caption{\label{all_setting}
AUC$_{norm}$ (\%) of all EIDI$_i$ settings.
}
\end{table}

\section{Frequency Bias}
\label{apd:c}
We also tested our pipeline on the frequency bias using the same dataset split measure as that for attestation bias. The dataset that we use is from  \citet{mckenna-etal-2023-sources}'s work, where we have 972 entries of frequency-consistent input and 220 entries of frequency-adversarial input.

Compared to baselines, the EIDI pipeline introduces extra frequency bias. This is expected since our transformation module is not designed to alter the relative frequency of the predicates, and may have amplified the frequency bias by taking popular alternative entities generated by the LLMs.
This result reaffirms the challenging nature of directional inference and the difficulty in improving robustness against multiple biases at once.

\begin{table}[h]
\centering
\begin{tabular}{@{}llccc@{}}
\toprule
\textbf{Model} & \textbf{Pipeline}                     & \textit{cons.} & \textit{adv.}              & \textit{\textbf{diff.}} \\ \midrule
GPT-3.5 & \multicolumn{1}{l|}{MCQ$_{entity}$}         & 20.58& \multicolumn{1}{c|}{29.38}  & +8.80\\
        & \multicolumn{1}{l|}{MCQ$_{type}$}          & 24.49& \multicolumn{1}{c|}{32.93}  & +8.44\\
               & \multicolumn{1}{l|}{EIDI$_{all}$} & 40.66& \multicolumn{1}{c|}{20.83}  & -19.83\\
        & \multicolumn{1}{l|}{EIDI$_1$} & 33.94& \multicolumn{1}{c|}{18.83}  & -15.11\\ \midrule
Llama3  & \multicolumn{1}{l|}{MCQ$_{entity}$}         & 33.30& \multicolumn{1}{c|}{47.87}  & +14.57\\
        & \multicolumn{1}{l|}{MCQ$_{type}$}          & 31.47& \multicolumn{1}{c|}{47.19}  & +15.72\\
               & \multicolumn{1}{l|}{EIDI$_{all}$} & 51.97& \multicolumn{1}{c|}{42.27} & -9.70\\
        & \multicolumn{1}{l|}{EIDI$_1$} & 39.78& \multicolumn{1}{c|}{35.32} & -4.46\\ \bottomrule
\end{tabular}
\caption{
Normalized area-under-curve(\%) on the frequency-bias-split datasets. 
}
\vspace{-0.2in}
\label{tab:my-table}
\end{table}

\section{Computational Cost}

Our experiments on Llama3-70B-Instruct are applied on two A6000 GPUs. For 1784 entries and 10 alternative inquiries for each entry, the typing module takes about 3 GPU hour, the transformation module takes about 100 GPU hours, and the prediction module takes about 6 GPU hours.

\end{document}